\documentclass{article}

\usepackage{graphicx}
\usepackage{multirow}
\usepackage{amsmath,amssymb,amsfonts}
\usepackage{amsthm}
\usepackage{mathrsfs}
\usepackage[title]{appendix}
\usepackage{xcolor}
\usepackage{textcomp}
\usepackage{booktabs}
\usepackage{algorithm}
\usepackage{algorithmicx}
\usepackage{algpseudocode}
\usepackage{authblk}
\usepackage{listings}
\usepackage{tikz}
\usetikzlibrary{positioning,shapes.geometric}
\usepackage[left=0.9in,right=0.9in,bottom=1.0in]{geometry}
\usepackage[numbers,sort&compress]{natbib}

\usepackage[
  colorlinks=true,
  citecolor=blue,
  linkcolor=blue,
  urlcolor=blue
]{hyperref}

\theoremstyle{plain}

\theoremstyle{definition}

\theoremstyle{remark}

\begin{document}

\title{\textit{Evaluating Large Language Models (LLMs) in Financial NLP: A Comparative Pilot Study on Financial Report Analysis}}

\author{Md Talha Mohsin\thanks{Corresponding author. Email: mdtalhamohsin@utulsa.edu}}

\affil{University of Tulsa,  
800 S Tucker Dr, Tulsa, OK 74104, USA}

\date{}

\maketitle

\begin{abstract}
Large language models (LLMs) are increasingly used to support the analysis of complex financial disclosures, yet their reliability, behavioral consistency, and transparency remain insufficiently understood in high-stakes settings. This paper presents a controlled evaluation of five transformer-based LLMs applied to question answering over the Business sections of U.S. 10-K filings. To capture complementary aspects of model behavior, we combine human evaluation, automated similarity metrics, and behavioral diagnostics under standardized and context-controlled prompting conditions. Human assessments indicate that models differ in their average performance across qualitative dimensions such as relevance, completeness, clarity, conciseness, and factual accuracy, though inter-rater agreement is modest, reflecting the subjective nature of these criteria. Automated metrics reveal systematic differences in lexical overlap and semantic similarity across models, while behavioral diagnostics highlight variation in response stability and cross-prompt alignment. Importantly, no single model consistently dominates across all evaluation perspectives. Together, these findings suggest that apparent performance differences should be interpreted as relative tendencies under the tested conditions rather than definitive indicators of general reliability. The results underscore the need for evaluation frameworks that account for human disagreement, behavioral variability, and interpretability when deploying LLMs in financially consequential applications.

\end{abstract}


\section{Introduction}

Artificial Intelligence (AI) has rapidly transformed how information is processed, analyzed, and interpreted across a multitude of industries. One of its biggest advances has been the development of LLMs, a category of AI systems that are trained on large text corpora to produce human-like text. In the recent past, LLMs have attracted a lot of attention in the field of academia and business \cite{bommasani_opportunities_2022}, \cite{zhao_survey_2023}, \cite{wei_emergent_2022}. These models are being used more and more in professional fields for their effectiveness, accuracy and rapidness.

With the rise in the need to have scalable, low-cost tools to assess strategies, LLMs are engaged in interpreting financial disclosures, sentiment analysis in the market, and aggregating unstructured data \cite{daimi_framework_2024}. For example, \cite{kim_financial_2025} states how LLMs can work wonders because they can predict changes in company earnings better than human analysts can, even when there aren't any clear narratives or industry-specific signals. They can also show how people feel about a company and any possible biases that might affect how investors act and how prices change \cite{nakagawa_bias_2024}. Additionally, LLMs can create financial digests by condensing large amounts of text into clear, actionable insights \cite{lazarev_trend_2024}; this leads to better trading methods with higher Sharpe ratios \cite{kim_radiology_2025}. In these situations, it is very important to know how to use domain-specific language and make products that can be understood. Because of this, LLMs are now judged not only on their fluency or coherence, but also on their organized, expert-level analysis in complex Financial Natural Language Processing (FinNLP) tasks.

A lot of natural language processing tasks, like summarizing, answering questions, and semantic document analysis, have been done very well by LLMs. But not much is known about how different LLMs act in high-stakes, domain-specific situations like company financial disclosures. This is especially worrying in the sense that in the financial world, regulators, institutional investors, and analysts depend on complicated textual disclosures heaivily to make decisions. Transformer-based LLMs, which are pre-trained on huge datasets and fine-tuned to spot complex language patterns \cite{kim_radiology_2025}, have sped up recent progress in natural language processing (NLP). In finance, NLP methods have mostly been used for classifying emotions or recognizing entities. However, new research is looking into how well they can be used for semantic reasoning and understanding stories \cite{wu_bloomberggpt_2024}. Since financial reports are mostly text, it makes sense to use AI-powered tools to analyze them \cite{abdaljalil_summarization_2021}.

The Securities and Exchange Commission (SEC) requires public companies to file structured textual disclosures, like the 10-K annual report. Important soft data that quantitative predictors miss can be found in these filings \cite{lombardo_format_2024}. The evaluation of extensive documents, in particular 10-K, could give crucial information about the work of a company. The filings are very comprehensive, and thus they are full of qualitative information. They are very useful in analyzing strategy, risk exposure, competitor positioning, and others, particularly when processed with a natural language processing. The 10-K filing set includes financial, risk, and strategy reports, which provide a brief but still systematic perspective on the well-being of a corporation. However, such documents are not so well-known in their comprehension. The format and the use of technical jargon is a challenge to both human analysts and automated systems as it is difficult to extract any meaningful information. The text mining provides a compelling method of extracting value out of this disclosure, highlighting patterns, sentiments and relationship details between businesses as well as industries \cite{kim_sec_sentiment_2023}.

This study address that gap by suggesting a multi-dimensional evaluation approach to see how five cutting-edge LLMs— ChatGPT-4, Perplexity, Claude 4 Opus, Gemini, and DeepSeek—look at the Business section (Item 1) of 10-K filings from the "Magnificent 7" tech companies over the last three years. Focusing on large, information-rich disclosures from major technology firms allows for a controlled and demanding evaluation setting, but the analysis is intended to capture relative behavioral tendencies under standardized conditions rather than to serve as a comprehensive benchmark of financial reporting more broadly. We introduce a Chain-of-Thought (CoT) prompting approach that tells models to behave like financial analysts in order to simulate real-life analytical workflows. Their reactions are assessed on the basis of three different aspects: human-based evaluation, computerized metric based grading, and immediate sensitivity behavior evaluation. Our work makes the following three contributions: (i) We test how sensitive LLM is to the design and amount of information in prompts, showing patterns of behavior across model structures. (ii) We create a benchmark that can be replicated to test LLMs' understanding of financial matters by focusing on detailed information rather than numbers, and (iii) We give academics and professionals a direction on how to use or evaluate LLMs in financial situations, mainly for strategic analysis, and information extraction The remainder of this study is structured as follows: Section 2 discusses background, Section 3 provides contextual insights into how LLMs work, Section 4 outlines the data and methodology, Section 5 presents the findings, section 6 offers discussion, and Section 7 concludes the study.

\section{Related Work}\label{sec2}

\subsection{Financial Text Analysis Using NLP}

The systematic, and consistent quality of content analysis has made it a fundamental component of qualitative research for a long time. This is especially true in the field of finance, where content or text analysis has undergone substantial development since the introduction of Natural Language Processing (NLP). By mining massive amounts of financial text, modern natural language processing  systems are now able to assist educated investment strategies, track macroeconomic signals, and improve decision-making processes within institutions \cite{textual_finance_ar_2025}. When new natural language processing (NLP) and information retrieval technologies come into existence, the meaning of "new" information and the accessibility of it evolved, providing early adopters with a temporary advantage \cite{araci_finbert_2019}.  These techniques integrate linguistic, statistical, and deep learning methodologies in order to extract value from complicated documents \cite{wang_nlp_risk_2024}. Natural language processing (NLP) is a strong tool in automated extraction of structure of unstructured financial text. \cite{oyewole_automation_2024}  state that the techniques are becoming more widespread. They enhance more accurate reporting, facilitate regulatory compliance and contribute to more efficient risk indicators identification.

Before NLP, Word2Vec, GloVe, and FastText were some of the first embedding models that learned word associations based on co-occurrence patterns. However, these models lacked contextual complexity in learning vocabulary. Regardless of the contexts in which a word is used, a single static vector was allocated to each individual word. Since then, this constraint has been overcome thanks to the development of contextualized embeddings through transformer-based models. These models encode words in a dynamic manner while taking into account the text that is surrounding them \cite{huang_finbert_2023}.

In recent years, the diverse use of FinNLP systems has entered the financial domain. Word embeddings and sentiment analysis technologies have allowed these systems to be more accurate and scalable to determine financial risk and understand the stories they report on \cite{wang_nlp_risk_2024}, \cite{sehrawat_embeddings_2019}. These NLP techniques enable analysts to have real-time information about past unobservable sentiment and theme changes in large volumes of data \cite{zhou_conversational_2024}.

\subsection{Emergence of LLMs in Financial NLP}

A Large Language Model (LLM) is an AI algorithm that uses deep learning and large datasets to interpret, summarize, combine, and predict new information \cite{prompt_engineering_2024}. LLMs use transfer learning: they are initially trained on large text datasets to generalize language abilities and then trained on specific tasks with impressive results \cite{luo_pretrained_2024}, \cite{zhang_retrieval_2023}. LLMs are used to process the incoming text by transforming into high-dimensional vectors. They do this through multilayer transformer networks that are able to capture the meaning and contextual relationships among words and phrases. Responses are made by predicting the next token in an autoregressive way, using the statistical patterns that have been learned. The quality of these answers depends on a number of things, such as the input prompt, which affects the context and specificity; the model's hyperparameters, which control how it makes inferences; and the variety of the training data, which determines how much knowledge the model has \cite{bender_dangers_2021}.  LLMs use probabilistic token selection methods to create outputs, which means that the same inputs and prompts might lead to different outputs in different runs \cite{wang_consistency_2025}. It is slowly replacing a number of other models on numerous NLP tasks when it comes to large language models. They do this through learning by means of large amounts of training data and by uncovering valuable patterns in financial data that they have never seen before \cite{zhang_retrieval_2023}.

\begin{figure}
    \centering
    \includegraphics[width=1\linewidth]{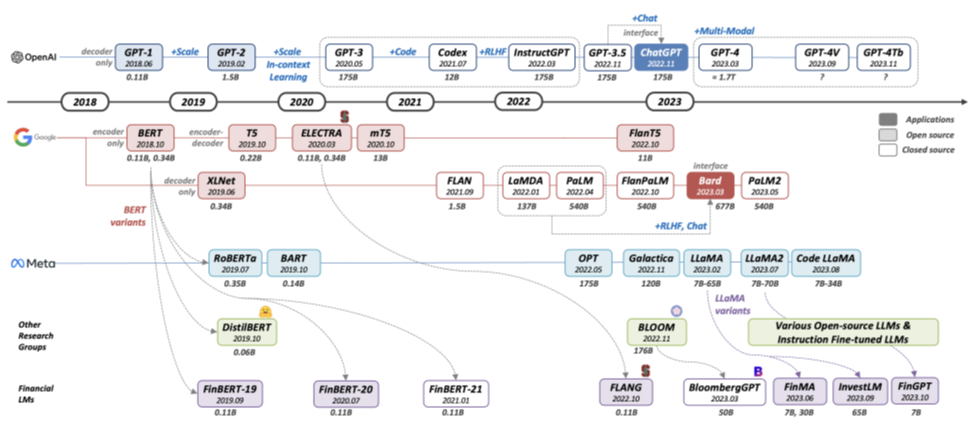}
    \caption{Timeline showing the evolution of selected PLM/LLM releases from the general domain to the financial domain \cite{lee_survey_finllms_2024}.}
    \label{fig:enter-label}
\end{figure}

OpenAI's GPT series and other LLMs have made a lot of progress in natural language processing (NLP) in the last several years. The progress of these models marks a major step forward for AI to understand and create natural language. LLMs have kept up with numerous language generation and comprehension challenges successfully; specially they are highly promising in the field of finance \cite{chang_evaluation_2024},
\cite{wu_chatgpt_2023}, \cite{wu_bloomberggpt_2024}. Due to their ease of use, affordability, speed, and capability to perform numerous text-analysis operations, such as annotation, classification, sentiment analysis, and critical discourse analysis, academics think that LLMs will transform text analysis \cite{tornberg_text_2023}.

Financial LLMs (Fin-LLMs) are a research subfield, as a result of the fast progress of general-domain LLMs. They use models based on mixed-domain LLMs that are fast to engineer and can be improved by prompt engineering \cite{lee_survey_finllms_2024}. They may be adapted to categorize headlines on financial news \cite{luo_pretrained_2024} and anticipate future financial statements and earnings \cite{kim_financial_2025}, among other specialized NLP tasks, such as those in financial statements and forecasting income in the future \cite{araci_finbert_2019}. \cite{yu_fincon_2024} present FINCON, a multi-agent framework based on LLM that uses a manager-analyst hierarchy to help agents work together across functions to reach common goals through natural language interactions for a range of financial tasks. \cite{wang_consistency_2025} focuses on building an intelligent financial data analysis system using LLM-RAG for financial analysis tasks, which provides important information for future improvements in intelligent financial data processing systems. FinBen, on the other and, is more than just a benchmark; it's a research platform that offers new challenges, datasets, evaluation methods, and ways for the community to participate to push the boundaries of financial LLM development \cite{xie_finben_2024}. In \cite{wang_fingpt_2023}, the authors introduce FinGPT, an open-source project of the financial large language models (FinLLM). They stress that the creation of open-source FinLLMs should be based on an extensive data gathering, cleaning, and preparation. \cite{yu_finmem_2024} introduce FINMEM, the first LLM-based autonomous trading agent with a novel layered memory architecture and adaptable character design. Unlike previous LLM agents in finance, FINMEM has a memory module that can handle financial data from several sources. \cite{li_investorbench_2024} shows how INVESTORBENCH, a new and broad financial benchmark, tests the reasoning and sequential decision-making skills of LLM-based agents in complex, open-ended financial situations.

\subsection{Prompt Design and Sensitivity in LLMs}

As LLMs get better, it's important to understand how sensitive they are to prompts in order to make sure they work well and reliably \cite{anagnostidis_prompt_2024}. Some of the current LLMs often don't do well in the financial segment since there are big variations between general text data and financial text data \cite{wang_fingpt_2023}. Also, LLMs usually agree with what the users think, even if it's different from what they think, which shows a Clever Hans effect in LLMs \cite{anagnostidis_prompt_2024} and as a result, they often act like sycophants \cite{perez_model_written_2023}. For instance, \cite{binz_cognitive_2023} shows how GPT-3 does several amazing things: it does vignette-based tasks as well as or better than humans, it makes good decisions based on descriptions, it beats people in a multi-armed bandit test, and it shows signs of model-based reinforcement learning. At the same time, even small changes to vignette-based tasks can throw GPT-3 off track, since it shows no signs of deliberate inquiry and does badly on a causal reasoning challenge. So, it's really important to ask the right questions as the input prompts together with temperature settings affect how LLMs make decisions \cite{loya_sensitivity_2023}. 

Prompt engineering is the methodical grouping of inputs and has become a crucial strategy for boosting the efficacy and precision of LLM models (\cite{chen_prompt_2025}. The newest generation of LLMs may be pushed to do amazing things with zero-shot or few-shot performance in a lot of NLP tasks through prompt engineering \cite{leidinger_prompting_2023}. Small changes in prompt can go a long way to get the proper results. LLMs react to changes in prompts (task instructions, prompt structure, few-shot instances, and debiasing prompts) by looking at how well they do on tasks and how biased they are in social situations \cite{hida_social_2024}. So it is extremely important to notice how LLMs are quite sensitive to how prompts are written, which can have a big impact on how well they can give the right answers \cite{benchmarking_prompt_2025}.

Existing financial language models such as FinBERT \cite{araci_finbert_2019}, BloombergGPT \cite{wu_bloomberggpt_2024}, and FinGPT \cite{wang_fingpt_2023} focus on domain-specific pretraining and supervised tuning on financial sentiment, news classification, and event extraction tasks. While these models establish strong baselines for token-level or sentence-level prediction, they do not evaluate cross-model consistency, reasoning stability, or prompt sensitivity when answering multi-sentence analytical questions grounded in regulatory disclosures. Prior benchmarks also rely almost exclusively on automated quantitative metrics, without incorporating human evaluators to assess factual coherence, interpretive accuracy, or domain-relevant reasoning. In contrast, our study integrates expert human judgments with semantic and lexical similarity metrics, enabling a more robust assessment of model reliability. Moreover, existing benchmarks concentrate on short-text tasks, whereas we examine long-form reasoning over full 10-K Business sections—a setting for which no standardized human–LLM evaluation framework currently exists. These differences highlight a gap in financial NLP research: the absence of a systematic, human-validated, multi-model comparison of interpretability, semantic alignment, and stability in high-stakes disclosure analysis.

\section{Language Modeling in Transformer-Based LLMs}\label{sec2}

Let a sequence of discrete tokens be denoted by
\[
W = (w_1, w_2, \ldots, w_n), \qquad w_i \in \mathcal{V},
\]
where $\mathcal{V}$ is a finite vocabulary. The objective of a language model is to learn a probability measure $P_\theta$ over the space of all finite-length token sequences $\mathcal{V}^*$ such that the induced distribution approximates the unknown data-generating process.

\subsection{Autoregressive Factorization}

By the chain rule of probability, the joint distribution over $W$ can be decomposed as
\[
P_\theta(W) = \prod_{i=1}^{n} P_\theta(w_i \mid w_{<i}),
\]
where $w_{<i} = (w_1,\ldots,w_{i-1})$. This autoregressive factorization converts sequence modeling into a sequence of conditional prediction problems.

Classical statistical language models approximate each conditional probability using a finite-order Markov assumption:
\[
P(w_i \mid w_{<i}) \approx P(w_i \mid w_{i-(k-1)},\ldots,w_{i-1}),
\]
for small $k$. While computationally tractable, such models suffer from an exponential growth in parameter space and are fundamentally incapable of representing long-range dependencies.

\subsection{Parametric Representation via Transformers}

Transformer-based LLMs replace fixed-order approximations with a parametric mapping from the entire prefix $w_{<i}$ to a continuous latent space. Let $\mathbf{x}_j \in \mathbb{R}^d$ denote the embedding of token $w_j$. The model computes a contextual representation
\[
\mathbf{h}_i = f_\theta(\mathbf{x}_1,\ldots,\mathbf{x}_{i-1}),
\]
where $f_\theta$ is implemented as a composition of multi-head self-attention and position-wise feedforward layers.

Formally, a single self-attention head computes
\[
\mathrm{Attn}(Q,K,V) = \mathrm{softmax}\!\left(\frac{QK^\top}{\sqrt{d_k}}\right)V,
\]
where $Q = XW_Q$, $K = XW_K$, and $V = XW_V$ are learned linear projections of the input embeddings. Multi-head attention aggregates multiple such heads to capture heterogeneous dependency patterns across positions.

Stacking $L$ such layers yields a sequence of transformations
\[
\mathbf{H}^{(\ell)} = \Phi^{(\ell)}(\mathbf{H}^{(\ell-1)}), \qquad \ell = 1,\ldots,L,
\]
with $\mathbf{H}^{(0)} = [\mathbf{x}_1,\ldots,\mathbf{x}_{i-1}]$, producing the final hidden state $\mathbf{h}_i$ used for prediction.

\subsection{Conditional Distribution and Output Layer}

The latent representation $\mathbf{h}_i$ is mapped to the vocabulary space via an affine transformation
\[
\mathbf{z}_i = W_o \mathbf{h}_i + \mathbf{b}_o,
\]
where $W_o \in \mathbb{R}^{|\mathcal{V}| \times d}$ and $\mathbf{b}_o \in \mathbb{R}^{|\mathcal{V}|}$.

The conditional probability of emitting token $v \in \mathcal{V}$ is given by the softmax distribution
\[
P_\theta(w_i = v \mid w_{<i}) =
\frac{\exp(z_{i,v})}{\sum_{v' \in \mathcal{V}} \exp(z_{i,v'})}.
\]
This formulation ensures that $P_\theta(\cdot \mid w_{<i})$ is a valid categorical distribution over the vocabulary.

\subsection{Training Objective and Statistical Interpretation}

Model parameters $\theta$ are estimated by minimizing the empirical negative log-likelihood over a corpus $\mathcal{D}$:
\[
\mathcal{L}(\theta)
= - \mathbb{E}_{W \sim \mathcal{D}}
\left[
\sum_{i=1}^{|W|} \log P_\theta(w_i \mid w_{<i})
\right].
\]
Equivalently, this objective minimizes the cross-entropy between the empirical data distribution and the model distribution. Under standard assumptions, this is equivalent to minimizing the Kullback–Leibler divergence
\[
\mathrm{KL}(P_{\text{data}} \,\|\, P_\theta),
\]
up to an additive constant independent of $\theta$.

\subsection{Decoding, Approximation, and Stochasticity}

At inference time, generation corresponds to sequential sampling from the learned conditional distributions:
\[
w_i \sim P_\theta(\cdot \mid w_{<i}),
\]
often modified via temperature scaling, nucleus sampling, or beam search. These decoding procedures introduce additional stochasticity and approximation error, causing realized outputs to deviate from the maximum-likelihood sequence even when $\theta$ is fixed.

Consequently, observed model responses should be interpreted as samples from an implicit conditional distribution rather than deterministic solutions to an optimization problem.

\subsection{Implications for Evaluation}

All the language models that we have studied, GPT, Perplexity, Claude, Gemini, and DeepSeek, are all based on the same probabilistic framework. Their key distinctions are scale, training data and optimization strategies. Subsequently, the models are examined by examining the actual model outputs, which indicate the probability distribution of the model, i.e., it is given by $P_\theta(w_i \mid w_{<i})$ rather than direct access to $\theta$.

We also have human reviewers to determine the appearance of the sampled sequences. There are automated similarity scores which estimate the distance between the generated distributions. In behavioral diagnostics, the stability and consistency of models across a variety of samples are measured. These approaches combined provide us with indirect but comprehensive information on how well each of those models is approximated by their actual outputs to the theoretical conditional distributions that we specified above.

\section{Experimental Design} \label{sec2}

\subsection{Data}

We selected five transformer-based LLMs, GPT-4, Claude 4 Opus, Gemini Pro, Perplexity, and DeepSeek for evaluation. These models are applied to the Item 1 (Business) section of annual 10-K filings from seven U.S. technology firms commonly referred to as the “Magnificent 7” companies. Our methodology follows a set of clearly defined phases, including corpus construction, prompt design, model execution, and multi-faceted evaluation.

The companies are: Apple, Microsoft, Amazon, Alphabet, Nvidia, Meta, and Tesla; each file an annual 10-K report that includes Item~1:~Business, a narrative overview of operations, strategic positioning, and market conditions. For this study, we extracted the Item~1 sections from the 2022, 2023, and 2024 filings, yielding 21 documents. Because these firms do not share a common fiscal year end, their 10-K filings become available at different times. In particular, Nvidia’s fiscal year closes in late January, so its ``2025'' filing was released earlier than the 2025 filings of the other companies; functionally, this filing is temporally aligned with the 2024 disclosures of the remaining firms. Text was obtained from the SEC’s EDGAR system, converted from HTML, and manually preprocessed to remove boilerplate language and ensure consistent formatting across companies and models. Because these firms are exceptionally large and produce unusually detailed disclosures, the resulting sample is not representative of the broader reporting landscape. Accordingly, the analysis should be interpreted as an exploratory evaluation rather than a comprehensive benchmark, and future work will require larger and more diverse sets of firms, industries, and disclosure types to support broader generalizability.

\subsection{Prompt Engineering and Tasks Designing}

We created a series of 10 open-ended interpretative questions to test LLMs' capacity to extract, combine, deduce, and interpret financial information in Table \ref{tab:company_questions}. We kept improving the prompts until we found a good mix between how easy they were to understand, how well they could be used in different situations, and how relevant they were to the subject. The last series of questions covers things like strategic intent, business model reasoning, risk inference, stakeholder framing, and looking ahead to the future. Each question is meant to get analytical answers instead of just extractive summaries, so the models have to make conclusions or find hidden patterns. Each company-year document got its own set of prompts, so there were no memory artifacts or context carryover. To keep previous conversations from leaking into new ones, prompting was done in fresh, separate chat rooms for each model-document pair.

\definecolor{lightbluebox}{RGB}{240,248,255}
\definecolor{blueframe}{RGB}{140,170,210}

\begin{center}
\begin{tikzpicture}
\node[
    draw=blueframe,
    fill=lightbluebox,
    rounded corners=8pt,
    inner sep=12pt,
    text width=0.95\linewidth
] (box) {
    \small
    \setlength{\parskip}{4pt}
    \setlength{\parindent}{0pt}
    \begin{enumerate}
        \item What are the company’s indicated strategic goals for the next two to three years?
        \item What is the company’s competitive position, and what evidence supports this assessment?
        \item What is the business description about the growth strategy of the company?
        \item Do your operational problems as not explicitly defined?
        \item What do you think is the business strategy that fits the company the most? Why?
        \item What are the number of the different lines of business that the corporation runs and is it over-reliant on a single line?
        \item What is the company’s stated value proposition, and who are its primary stakeholders?
        \item What key points should a stakeholder or investor remember about this company?
        \item On a scale of 1 to 5, how clearly does the company explain its business model? Justify the rating.
        \item How forward-looking is the business description?   Does it emphasize future plans or mainly summarize current operations?
    \end{enumerate}
};
\end{tikzpicture}
\end{center}

\subsection{Experimental Configuration}

This section describes the full experimental setup used in evaluating the five proprietary LLMs: GPT\textendash 4 (OpenAI), Claude~4 Opus (Anthropic), Gemini Pro (Google DeepMind), Perplexity (Perplexity AI), and DeepSeek\textendash V2. All systems are transformer-based autoregressive decoders, but they differ in model scale, training corpora, alignment procedures, and interface constraints. To ensure transparency and reproducibility, we report all prompt formats, sampling settings, token limits, and chain-of-thought (CoT) configurations used in the study.

\subsubsection{Model Access and Decoding Parameters}

All models were accessed through their official web interfaces between August and October 2025. When decoding controls were exposed (GPT\textendash 4 and Perplexity), we used deterministic sampling with temperature set to $0$ and top-$p$ fixed at $1.0$. Claude and Gemini do not permit user-adjustable decoding parameters; therefore, their default sampling settings were used. DeepSeek\textendash V2 was run in its standard deterministic mode. No beam search, reranking, multi-sample decoding, or self-consistency sampling was employed. Each query produced exactly one response without retries or regeneration.

\subsubsection{Input Length and Token Constraints}

To ensure comparability across models, each 10-K \textit{Business} section was truncated to approximately $3{,}000$--$3{,}500$ tokens. This limit corresponds to the smallest effective context window among the evaluated systems, namely Perplexity and DeepSeek. The identical truncated input passages and question sets were provided to all models. For each query, both input and output token counts were recorded.

Decoding and interface configurations for all models are summarized in Table~\ref{tab:decoding}. When explicit decoding controls were available (e.g., temperature and top-$p$), they were set to deterministic or near-deterministic values to minimize stochastic variation. For models exposed only through fixed web interfaces, default decoding settings were used.

\begin{table}[h!]
\centering
\caption{Decoding and interface settings for all evaluated models.}
\label{tab:decoding}
\begin{tabular}{lccccc}
\toprule
\textbf{Model} & \textbf{Temperature} & \textbf{Top-$p$} & \textbf{Max Tokens} & \textbf{Interface} & \textbf{Sampling Mode} \\
\midrule
GPT-4 (OpenAI)        & 0.0 & 1.0 & $\sim$4{,}096 & Web UI & Deterministic \\
Claude 4 Opus        & Default ($\approx0.6$) & Default & $\sim$4{,}000 & Web UI & Interface-default \\
Gemini Pro           & Default ($\approx0.7$) & Default & $\sim$8{,}000 & Web UI & Interface-default \\
Perplexity (PPLX)    & 0.0 & 1.0 & $\sim$4{,}000 & Web UI & Deterministic \\
DeepSeek-V2          & Deterministic & N/A & $\sim$4{,}000 & Web UI & Deterministic \\
\bottomrule
\end{tabular}
\end{table}

\subsubsection{Prompt Template and Example}

All models received the same instruction structure. The base prompt used was:

\begin{quote}
\small
\texttt{You are analyzing the Business section of a firm's 10-K filing.\\
Use only the information contained in the passage. Do not introduce external knowledge.}

\texttt{[10-K Business Section Excerpt]}

\texttt{Question: [Q\_i]\\
Provide a concise, factual answer.}
\end{quote}

No additional system prompts, memory functions, or contextual hints were used. A fully instantiated prompt, including an actual 10-K excerpt and question, is provided in Appendix~\ref{app:exampleprompt}.

\subsubsection{Chain-of-Thought Protocol}

Two prompting modes were evaluated.

\paragraph{Primary (Non-CoT) Mode.}
For the main analysis, models were instructed to provide direct answers without revealing intermediate reasoning. These responses were used for all human evaluation scores and for all quantitative similarity metrics (cosine similarity, ROUGE-L, and Jaccard).

\paragraph{Diagnostic CoT Mode.}
To characterize reasoning behavior, a secondary diagnostic run used a controlled chain-of-thought prompt:

\begin{quote}
\small
\texttt{Think step-by-step in 2--4 concise points. Base each step only on information contained in the passage. Do not use external knowledge or assumptions. After outlining these steps, provide a single, clearly stated final answer.}
\end{quote}

CoT traces were analyzed solely for reasoning structure, verbosity, and stability. They were not scored, and annotators did not see CoT output. Summary findings from the diagnostic CoT runs are reported in Section~\ref{sec:cotresults}.

\subsubsection{Output Logging and Version Control}

For each query, we logged the complete prompt, model output, input and output token counts, timestamp, and model version identifier (when available). No manual edits or post-processing were applied to any model output. Across 21 filings and 10 questions per filing, each model generated $210$ responses, resulting in a corpus of $1{,}050$ model outputs. Five independent annotators evaluated every response, yielding $5{,}250$ human ratings.

This configuration provides complete transparency regarding the evaluation pipeline and ensures that all results reported in this study can be reproduced under identical conditions.

\begin{figure}[H]
  \centering
  \includegraphics[width=0.5\textwidth]{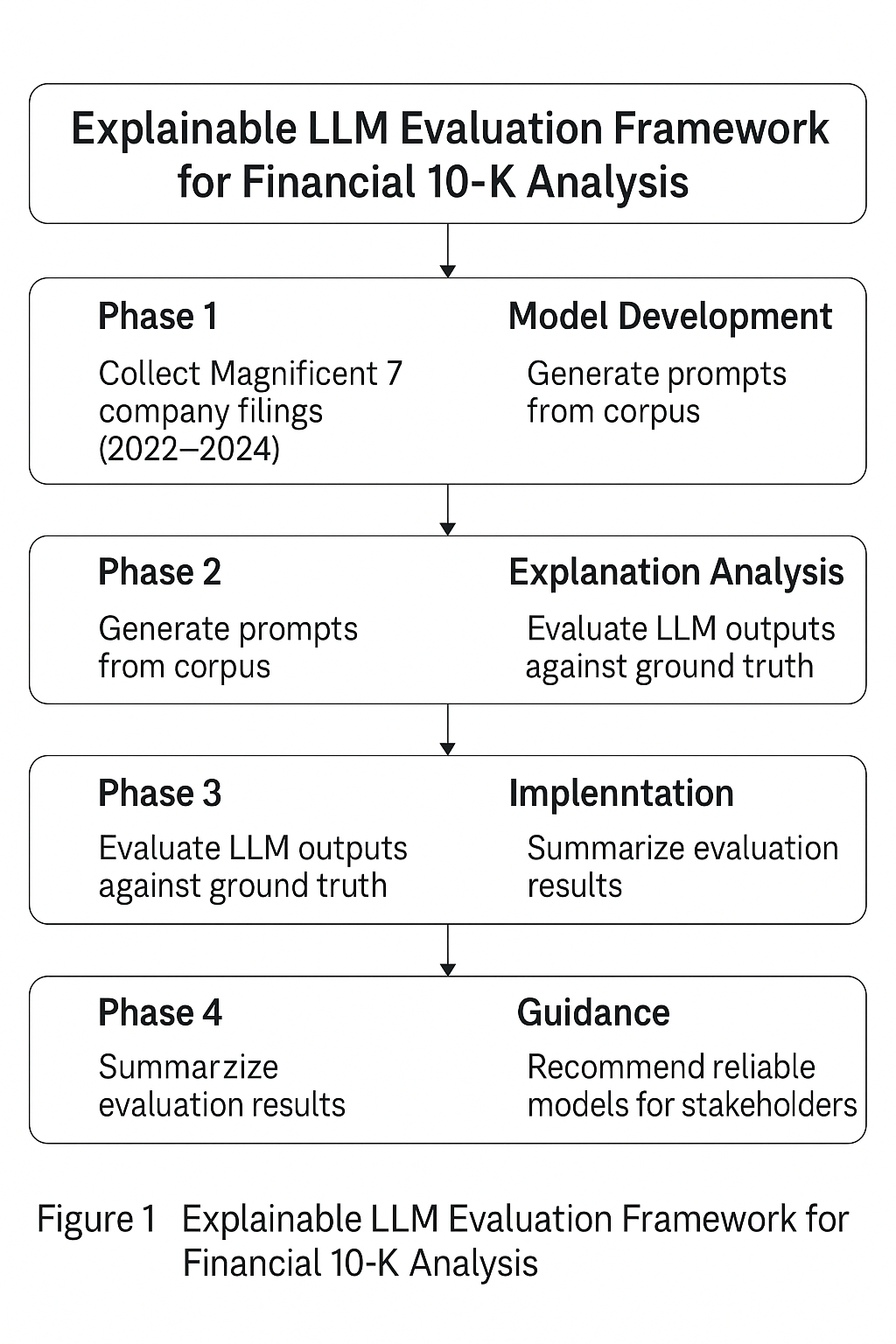}
  \caption{Overview of the explainable LLM evaluation framework used in this study.}
  \label{fig:framework}
\end{figure}

\subsection{Evaluation Framework and Scoring Metrics}

Let $m \in \mathcal{M}$ index models, $d \in \mathcal{D}$ documents, $q \in \mathcal{Q}$ prompts, and $a \in \mathcal{A}$ annotators.  
Each model $m$ produces a response $y_{m,d,q}$ for document--prompt pair $(d,q)$.  
Evaluation is conducted along three axes: human annotation, automated similarity metrics, and behavioral diagnostics.

\subsubsection{Human-Centric Evaluation}

Each response $y_{m,d,q}$ is independently rated by $|\mathcal{A}|=5$ annotators on five criteria:
\[
k \in \{\text{Relevance}, \text{Completeness}, \text{Clarity}, \text{Conciseness}, \text{Factual Accuracy}\}.
\]

Let $s_{a,k}(m,d,q) \in \{1,2,3,4,5\}$ denote annotator $a$’s score for criterion $k$.  
The mean human score for criterion $k$ is defined as
\[
\bar{s}_k(m) =
\frac{1}{|\mathcal{D}||\mathcal{Q}||\mathcal{A}|}
\sum_{d \in \mathcal{D}}
\sum_{q \in \mathcal{Q}}
\sum_{a \in \mathcal{A}}
s_{a,k}(m,d,q).
\]

An overall qualitative score for model $m$ is computed as
\[
\bar{s}(m) = \frac{1}{|K|} \sum_{k \in K} \bar{s}_k(m),
\]
where $K$ denotes the set of evaluation criteria.

The dataset consists of $21$ documents and $10$ prompts per document, yielding $210$ responses per model and $1{,}050$ responses in total. With five annotators, this produces $5{,}250$ individual ratings.

\subsubsection{Metric-Based Evaluation}

Automated metrics quantify lexical overlap and semantic alignment between model outputs and reference responses $y^{\ast}_{d,q}$.  

\paragraph{Lexical Similarity.}
ROUGE-$n$ and ROUGE-L measure $n$-gram and longest common subsequence overlap:
\[
\text{ROUGE}_n(y,y^{\ast}) =
\frac{\sum_{g \in \text{$n$-grams}(y^{\ast})} \min\{\text{count}_y(g), \text{count}_{y^{\ast}}(g)\}}
{\sum_{g \in \text{$n$-grams}(y^{\ast})} \text{count}_{y^{\ast}}(g)}.
\]

Jaccard similarity is defined as
\[
J(y,y^{\ast}) = \frac{|T(y) \cap T(y^{\ast})|}{|T(y) \cup T(y^{\ast})|},
\]
where $T(\cdot)$ denotes the set of unique tokens.

\paragraph{Semantic Similarity.}
Let $e(y) \in \mathbb{R}^d$ denote the Sentence-BERT embedding of response $y$.  
Cosine similarity is computed as
\[
\cos(y,y^{\ast}) =
\frac{e(y)^\top e(y^{\ast})}{\|e(y)\|_2 \|e(y^{\ast})\|_2}.
\]

Metric scores are averaged over documents and prompts:
\[
\overline{\cos}(m) =
\frac{1}{|\mathcal{D}||\mathcal{Q}|}
\sum_{d,q} \cos(y_{m,d,q}, y^{\ast}_{d,q}),
\]
with analogous definitions for ROUGE and Jaccard metrics.

\subsubsection{Model Behavior Diagnostics}

Cross-model agreement is measured via pairwise cosine similarity:
\[
A(m_i,m_j) =
\frac{1}{|\mathcal{D}||\mathcal{Q}|}
\sum_{d,q}
\cos\!\left(y_{m_i,d,q}, y_{m_j,d,q}\right).
\]

Prompt-level stability for model $m$ is quantified by the variance
\[
\mathrm{Var}_q(m) =
\frac{1}{|\mathcal{D}|}
\sum_{d}
\mathrm{Var}_{q}
\left(
\cos(y_{m,d,q}, y^{\ast}_{d,q})
\right),
\]
which captures sensitivity to prompt formulation.

Each prompt is executed in a different session in order to prevent contextual leakage. The responses with formatting errors, refusals, and hallucinations are flagged and are subject to qualitative analysis and removed from the metric totals.

\section{Results}\label{sec2}

We analyze the performance of LLMs in the context of providing high-quality, contextually accurate, and informative responses to questions generated based on the Business sections of 10-K filings in this section. We assess using the three major methods, namely human evaluation, metric based evaluation, and behavioral evaluation.

\subsection{Human Annotator Evaluation}
\subsubsection{Inter-Rater Reliability}
\label{subsubsec:inter_rater_reliability}

To assess the reliability of human evaluations, we measured inter-rater agreement across the five expert annotators for each LLM and evaluation dimension. As the ratings were ordinal and incomplete evaluations were permitted, we adopt Krippendorff’s $\alpha$ with an ordinal distance metric as the primary measure of inter-rater reliability. Krippendorff’s $\alpha$ accommodates multiple annotators, accounts for chance agreement, and naturally handles missing ratings. As a robustness check, we additionally report two-way random-effects intraclass correlation coefficients with absolute agreement (ICC(2,1)), which treat ratings as continuous and quantify the proportion of variance attributable to differences across evaluated texts.

\begin{table}[ht]
\centering
\caption{Inter-rater agreement measured by Krippendorff’s $\alpha$ (ordinal)}
\label{tab:alpha}
\begin{tabular}{lccccc}
\hline
\textbf{Dimension} & \textbf{Claude} & \textbf{DeepSeek} & \textbf{GPT} & \textbf{Gemini} & \textbf{Perplexity} \\
\hline
Clarity            & 0.033 & -0.051 & -0.065 & 0.079 & 0.111 \\
Completeness       & 0.016 & -0.071 & -0.001 & 0.077 & 0.061 \\
Conciseness        & 0.050 &  0.013 &  0.024 & 0.048 & 0.025 \\
Factual Accuracy   & 0.066 &  0.005 &  0.101 & -0.050 & 0.085 \\
Relevance          & 0.075 &  0.066 &  0.125 & 0.109 & 0.008 \\
\hline
\end{tabular}
\end{table}

\begin{table}[ht]
\centering
\caption{Inter-rater agreement measured by ICC(2,1) absolute agreement}
\label{tab:icc}
\begin{tabular}{lccccc}
\hline
\textbf{Dimension} & \textbf{Claude} & \textbf{DeepSeek} & \textbf{GPT} & \textbf{Gemini} & \textbf{Perplexity} \\
\hline
Clarity            & 0.024 & -0.056 & -0.070 & 0.099 & 0.115 \\
Completeness       & 0.013 & -0.079 & -0.008 & 0.076 & 0.060 \\
Conciseness        & 0.055 &  0.024 &  0.033 & 0.076 & 0.022 \\
Factual Accuracy   & 0.066 & -0.008 &  0.099 & -0.023 & 0.081 \\
Relevance          & 0.075 &  0.049 &  0.110 & 0.119 & 0.003 \\
\hline
\end{tabular}
\end{table}

Tables~\ref{tab:alpha} and~\ref{tab:icc} summarize inter-rater agreement by evaluation dimension and LLM. Agreement levels are generally low across all qualitative dimensions, with $\alpha$ values clustered near zero and occasional negative estimates. This pattern indicates that expert judgments frequently differ and, in some cases, diverge more than would be expected under chance agreement. ICC(2,1) values exhibit a closely aligned pattern, corroborating the conclusion that between-rater variability dominates between-item variability in this evaluation setting.

Agreement is comparatively higher for fact-anchored dimensions such as relevance and factual accuracy than for more interpretive criteria such as clarity and completeness, although reliability remains below conventional thresholds in all cases. These results are consistent with prior findings in expert evaluation of generated text, where qualitative dimensions lack a single objectively correct reference and evaluators apply heterogeneous internal standards.

Given the observed level of disagreement, we do not interpret individual annotator scores as ground truth. Instead, subsequent analyses rely on aggregated ratings and analytical models that explicitly account for rater-level variability. This approach ensures that reported differences across LLMs reflect systematic trends in human judgments rather than idiosyncratic evaluator preferences. Also, in Table \ref{tab:human-evaluation-metrics}, we can see the average scores that annotators gave each model on all five dimensions:

\begin{table}[ht]
\centering
\caption{Human Evaluation Scores Across LLMs}
\label{tab:human-evaluation-metrics}
\begin{tabular}{llccccc}
\toprule
\textit{Rater} & \textbf{Criteria} & \textbf{GPT} & \textbf{Perplexity} & \textbf{Claude} & \textbf{Gemini} & \textbf{DeepSeek} \\
\midrule
\multirow{5}{*}{\textbf{R1}} 
 & Relevance        & 4.14 & 3.90 & 4.10 & 3.95 & 3.57 \\
 & Completeness     & 4.00 & 3.95 & 3.81 & 4.00 & 3.76 \\
 & Clarity          & 3.95 & 3.90 & 4.05 & 4.05 & 3.90 \\
 & Conciseness      & 3.81 & 4.00 & 4.14 & 2.95 & 4.24 \\
 & Factual Accuracy & 4.00 & 3.95 & 4.19 & 4.43 & 3.71 \\
\midrule
\multirow{5}{*}{\textbf{R2}} 
 & Relevance        & 4.10 & 3.95 & 3.95 & 4.33 & 3.57 \\
 & Completeness     & 4.24 & 4.29 & 3.86 & 4.29 & 3.90 \\
 & Clarity          & 4.24 & 4.24 & 4.00 & 4.21 & 3.86 \\
 & Conciseness      & 4.00 & 3.95 & 4.00 & 3.45 & 4.14 \\
 & Factual Accuracy & 4.05 & 3.95 & 4.14 & 4.24 & 3.81 \\
\midrule
\multirow{5}{*}{\textbf{R3}} 
 & Relevance        & 4.14 & 3.90 & 4.19 & 4.52 & 3.67 \\
 & Completeness     & 4.05 & 3.95 & 4.10 & 4.43 & 3.86 \\
 & Clarity          & 4.24 & 4.10 & 3.95 & 4.10 & 4.19 \\
 & Conciseness      & 4.00 & 4.21 & 3.86 & 3.24 & 4.52 \\
 & Factual Accuracy & 4.10 & 4.10 & 4.19 & 4.57 & 4.00 \\
\midrule
\multirow{5}{*}{\textbf{R4}} 
 & Relevance        & 4.05 & 4.00 & 4.10 & 4.38 & 3.57 \\
 & Completeness     & 4.29 & 4.29 & 3.86 & 4.14 & 3.71 \\
 & Clarity          & 4.14 & 4.14 & 3.76 & 4.29 & 4.00 \\
 & Conciseness      & 3.95 & 3.90 & 3.95 & 3.57 & 4.19 \\
 & Factual Accuracy & 3.86 & 3.76 & 4.00 & 3.86 & 3.81 \\
\midrule
\multirow{5}{*}{\textbf{R5}} 
 & Relevance        & 4.10 & 3.95 & 3.90 & 3.86 & 3.86 \\
 & Completeness     & 4.00 & 4.05 & 4.00 & 4.05 & 3.95 \\
 & Clarity          & 4.29 & 3.95 & 3.95 & 3.95 & 3.90 \\
 & Conciseness      & 4.43 & 4.00 & 3.67 & 3.52 & 4.43 \\
 & Factual Accuracy & 3.95 & 3.90 & 4.00 & 3.95 & 3.81 \\
\midrule
\multicolumn{2}{l}{\textbf{Average}} & \textbf{4.08} & \textbf{4.01} & \textbf{3.99} & \textbf{4.01} & \textbf{3.92} \\
\bottomrule
\end{tabular}
\end{table}

\begin{figure}[htbp]
    \centering
    \includegraphics[width=0.8\textwidth]{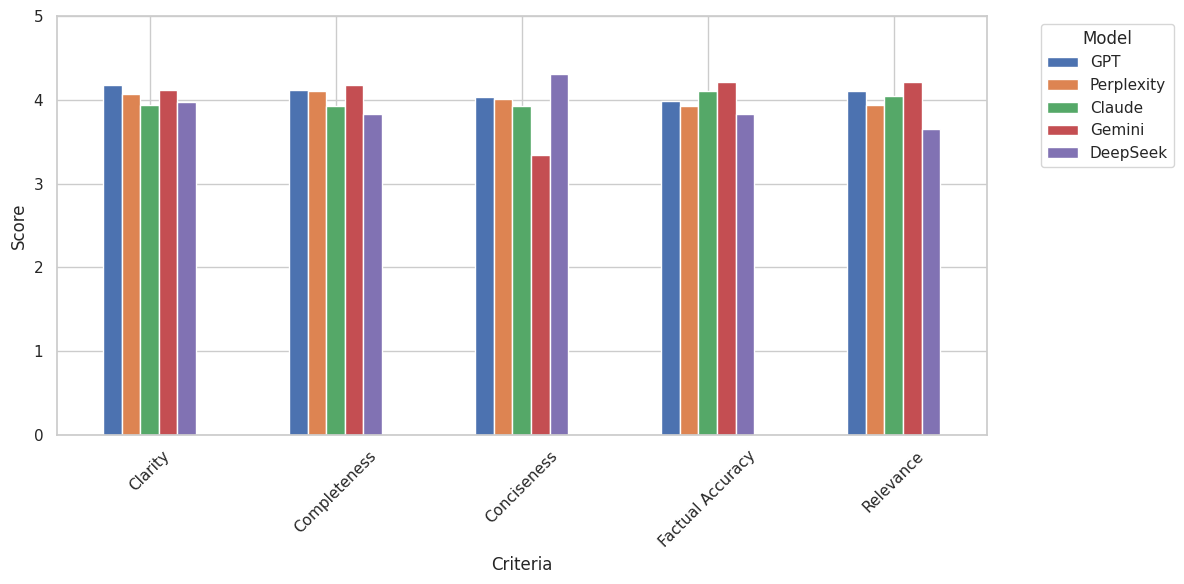}
    \caption{Human Evaluation Average Scores}
    \label{fig:human-eval-avg}
\end{figure}

Figure~\ref{fig:human-eval-avg} summarizes the average human evaluation scores across the five qualitative dimensions. GPT is rated by the highest number of annotators on the scale of relevance, completeness, and clarity. It also scores well on the accuracy of the facts. Claude compares especially in terms of factual accuracy, however, its clarity and completeness scores are a little lower than that of GPT. DeepSeek has shorter answers and is therefore more concise, but its relevance and factual accuracy score are lower. Perplexity is performing well on all dimensions. It lacks any significant strengths and weaknesses, but the factual accuracy of it is diminished to a certain extent by occasional factual inconsistencies. Gemini is less concise and factual, which is primarily due to the length and imprecision of its responses. However, its manifestation is quite comprehensible. On the whole, GPT and Claude perform better on average on the tested qualitative criteria. DeepSeek, Perplexity, and Gemini have different trade-offs with respect to response length, relevancy, and factual basis.

\subsection{ Metric Based Evaluation}

We used the following automatic evaluation metrics to get a fair look at the lexical and semantic quality of the model responses:
ROUGE (ROUGE-1, ROUGE-2, and ROUGE-L), Jaguar Similarity and Cosine Similarity (Sentence-BERT).

\begin{table}[ht]
\centering
\caption{Summary of Best Models and Their Scores by Metric}
\label{tab:metric_summary}
\begin{tabular}{l l c c}
\toprule
\textbf{Metric} & \textbf{Best Model} & \textbf{Avg Score} & \textbf{Range} \\
\midrule
ROUGE-1           & Gemini  & 0.56 & 0.20--0.62 \\
ROUGE-2           & Gemini  & 0.22 & 0.05--0.28 \\
ROUGE-L           & Gemini  & 0.16 & 0.06--0.17 \\
Cosine Similarity & Claude  & 0.68 & 0.44--0.80 \\
Jaccard           & Gemini  & 0.21 & 0.07--0.26 \\
\bottomrule
\end{tabular}
\end{table}

Table \ref{tab:metric_summary} shows the strengths of each model by measure. Gemini attains the highest ROUGE-1 score (0.56), indicating greater lexical overlap with reference responses under this metric. It looks like semantic tasks are more competitive because Claude has a smaller lead over GPT (0.68) and Perplexity (0.71) in Cosine similarity (0.68). The Jaccard scores (Gemini: 0.21) show that Gemini is very good with wordings, and the edge in ROUGE-2 shows how good Gemini is with different phrases. 

\begin{table}[ht]
\centering
\caption{Model Rankings Based on Average Metrics}
\label{tab:model_rankings}
\begin{tabular}{c l c c c c c}
\toprule
\textbf{Rank} & \textbf{Model} & \textbf{Avg R-1} & \textbf{Avg R-2} & \textbf{Avg R-L} & \textbf{Avg Cosine} & \textbf{Avg Jaccard} \\
\midrule
1 & Gemini     & 0.56 & 0.22 & 0.16 & 0.63 & 0.22 \\
2 & GPT        & 0.31 & 0.08 & 0.10 & 0.68 & 0.13 \\
3 & Perplexity & 0.29 & 0.08 & 0.10 & 0.71 & 0.13 \\
4 & Claude     & 0.27 & 0.08 & 0.09 & 0.67 & 0.12 \\
5 & DeepSeek   & 0.16 & 0.03 & 0.06 & 0.59 & 0.09 \\
\bottomrule
\end{tabular}
\end{table}

Table~\ref{tab:model_rankings} summarizes overall model performance relative to the reference responses using aggregate lexical and semantic similarity metrics. The evaluation findings demonstrate that models differ in results with regard to model metrics. Gemini ranks the highest on the measures based on ROUGE, that is, there is the highest lexical overlap with the reference texts. The other models, on the other hand, are more successful on semantic alignment measures. The values of cosine similarity between GPT and Perplexity (0.68 and 0.71 respectively) are similar in terms of their semantic proximity to the references. GPT also scores higher in ROUGE-1 (0.31), which means that it has wider lexical coverage. The metrics utilized by Claude are slightly less on a number of measures and DeepSeek less so on the measures of similarity considered. Such results demonstrate that there is a trade-off between the overlapping of lexical and semantic alignment across the models, as opposed to having a clear leader.

\begin{table}[ht]
\centering
\caption{Win Rates of Models Across Companies}
\label{tab:win_rates}
\begin{tabular}{l r r r r r}
\toprule
\textbf{Company} & \textbf{GPT} & \textbf{Perplexity} & \textbf{Claude} & \textbf{Gemini} & \textbf{DeepSeek} \\
\midrule
Amazon    & 58.60\% & 37.20\% & 25.10\% & 21.80\% & 19.20\% \\
Apple     & 59.10\% & 38.90\% & 24.20\% & 22.30\% & 19.60\% \\
Google    & 64.10\% & 40.10\% & 26.90\% & 20.90\% & 16.30\% \\
Meta      & 61.00\% & 36.30\% & 22.30\% & 18.40\% & 16.80\% \\
Microsoft & 66.40\% & 41.20\% & 30.50\% & 18.60\% & 15.90\% \\
Nvidia    & 62.50\% & 38.10\% & 23.90\% & 18.90\% & 14.70\% \\
Tesla     & 62.30\% & 36.60\% & 23.20\% & 20.50\% & 19.10\% \\
\bottomrule
\end{tabular}
\end{table}

Table~\ref{tab:win_rates} reports pairwise win rates from company-level comparisons across the evaluated LLMs. These win rates provide a relative summary of model performance under the behavioral evaluation protocol rather than a definitive ranking. GPT exhibits the highest average win rates across firms, ranging from approximately 58\% to 66\% depending on the company, with higher values observed for Microsoft-related prompts and lower values for Amazon-related prompts. Perplexity shows the next highest win rates, typically between 36\% and 41\%, indicating comparatively stronger performance among the remaining models. Claude achieves moderate win rates in the low to mid-20\% range, reflecting mixed outcomes across firms. Gemini and DeepSeek display lower win rates across all company-level comparisons, with values remaining below 22\%, suggesting more limited success under this evaluation setting. Overall, the win-rate patterns highlight relative performance differences across models while also indicating substantial overlap and variability across firms.

\subsection{Behavioral Diagnostics}

We examined Across-Model Cosine Similarity and Prompt-Level Response Variance to find out how consistent and generalizable each model is across different inquiries.

\begin{table}[ht]
\centering
\caption{Pairwise Similarity Scores Between Models Across Companies}
\label{tab:pairwise_similarity}
\begin{tabular}{lrrrrrrrrrr}
\toprule
\textbf{Company} & \textbf{G-P} & \textbf{G-C} & \textbf{G-G} & \textbf{G-D} & \textbf{P-C} & \textbf{P-G} & \textbf{P-D} & \textbf{C-G} & \textbf{C-D} & \textbf{G-Dk} \\
\midrule
Apple      & 0.77 & 0.84 & 0.77 & 0.84 & 0.76 & 0.69 & 0.77 & 0.79 & 0.84 & 0.78 \\
Amazon     & 0.80 & 0.82 & 0.79 & 0.80 & 0.78 & 0.72 & 0.71 & 0.71 & 0.77 & 0.77 \\
Alphabet   & 0.83 & 0.83 & 0.82 & 0.76 & 0.78 & 0.85 & 0.77 & 0.78 & 0.71 & 0.80 \\
Meta       & 0.78 & 0.84 & 0.82 & 0.76 & 0.76 & 0.79 & 0.75 & 0.75 & 0.79 & 0.75 \\
Microsoft  & 0.89 & 0.87 & 0.85 & 0.82 & 0.84 & 0.86 & 0.82 & 0.83 & 0.80 & 0.78 \\
Nvidia     & 0.81 & 0.78 & 0.81 & 0.77 & 0.88 & 0.86 & 0.73 & 0.84 & 0.72 & 0.72 \\
Tesla      & 0.81 & 0.79 & 0.79 & 0.78 & 0.76 & 0.79 & 0.75 & 0.74 & 0.80 & 0.75 \\
\bottomrule
\end{tabular}
\end{table}

Table \ref{tab:pairwise_similarity} shows mean Cosine Similarity of LLM Outputs by company. Using the same business-related text prompts extracted from company filings, this table compares the similarity between the outputs produced by various LLms. By calculating cosine similarity scores (ranging from 0 to 1) between all possible pairs of the models, it reveals how consistently these AI systems interpret and respond to the same input text across various companies.  It is easy to see trends in the averaged scores: some model combinations consistently provide comparable results (e.g., GPT and Claude have a high level of agreement with 0.84 similarity for Apple), while other pairs differ more noticeably (e.g., Perplexity and Gemini have a much closer relationship with Amazon at only 0.72 similarity). The resulting content can be significantly affected by the choice of LLM, even while processing the same source material, as these changes are most likely caused by differences in the models' training data, architectures, and algorithms.

\begin{table}[ht]
\centering
\caption{Cosine Similarity Scores and Variability Across Prompts}
\label{tab:cosine_similarity_prompts}
\begin{tabular}{lllrr}
\toprule
\textbf{Company} & \textbf{Year} & \textbf{Prompt ID} & \textbf{Mean Cosine} & \textbf{Std. Dev} \\
\midrule
Alphabet   & 2022 & Alphabet\_2022   & 0.778 & 0.070 \\
           & 2023 & Alphabet\_2023   & 0.787 & 0.041 \\
           & 2024 & Alphabet\_2024   & 0.812 & 0.034 \\
Amazon     & 2022 & Amazon\_2022     & 0.786 & 0.053 \\
           & 2023 & Amazon\_2023     & 0.807 & 0.050 \\
           & 2024 & Amazon\_2024     & 0.708 & 0.078 \\
Apple      & 2022 & Apple\_2022      & 0.767 & 0.063 \\
           & 2023 & Apple\_2023      & 0.774 & 0.055 \\
           & 2024 & Apple\_2024      & 0.820 & 0.035 \\
Meta       & 2022 & Meta\_2022       & 0.750 & 0.055 \\
           & 2023 & Meta\_2023       & 0.789 & 0.038 \\
           & 2024 & Meta\_2024       & 0.802 & 0.043 \\
Microsoft  & 2022 & Microsoft\_2022  & 0.846 & 0.029 \\
           & 2023 & Microsoft\_2023  & 0.851 & 0.042 \\
           & 2024 & Microsoft\_2024  & 0.810 & 0.041 \\
Nvidia     & 2023 & Nvidia\_2023     & 0.765 & 0.069 \\
           & 2024 & Nvidia\_2024     & 0.806 & 0.064 \\
           & 2025 & Nvidia\_2025     & 0.804 & 0.056 \\
Tesla      & 2022 & Tesla\_2022      & 0.751 & 0.062 \\
           & 2023 & Tesla\_2023      & 0.766 & 0.036 \\
           & 2024 & Tesla\_2024      & 0.813 & 0.038 \\
\bottomrule
\end{tabular}
\end{table}

Using mean cosine similarity and standard deviation, table \ref{tab:cosine_similarity_prompts} displays the consistency with which LLMs perceive identical business prompts across firms and years. The model agreement is highest for Microsoft (mean: 0.84-0.85, low standard deviation) and lowest for Amazon's 2024 prompt (mean: 0.71, high standard deviation: 0.078). Nvidia and Tesla exhibit greater volatility in prior years, whereas Apple and Alphabet display enhanced stability over time. The findings indicate that both the input data and the specific combinations of LLMs influence the extent of output similarity.

\section{Discussion}\label{sec13}

\subsection{Qualitative Financial Interpretation and Judgment}

This task evaluates how effectively each model performs qualitative financial interpretation under context-isolated conditions, focusing on relevance, completeness, clarity, conciseness, and factual correctness. Human annotations reveal systematic differences in how models balance interpretive depth, brevity, and factual grounding when analyzing 10-K disclosures.

GPT has the best total average of 4.08 on qualitative criteria, and in particular relevance, completeness, and clarity. It answers the source well and demonstrates consistency in facts, which indicates that it can absorb the financial context without introducing unsubstantiated data. Claude comes right behind with a mean rating of 3.99 and the top mean score in factual correctness (4.10). It is slightly less on conciseness and completeness, meaning that under assessed circumstances, Claude is focused on factual validity rather than brevity.

The perplexity is relatively stable by criteria with the majority of the dimensions having an average of slightly over 4.00 meaning that there is balanced and less differentiated behavior. DeepSeek has the highest score in conciseness (4.28), although it comes at the cost of reduced relevance (3.65) and factual accuracy (3.83). There are some cases where Gemini exhibits quite good syntactic structure and factual accuracy, but it has low conciseness (3.55) and high variability across raters, which leads to worse interpretive efficiency, particularly when verbosity cannot contribute valuable information.

\subsection{Lexical and Semantic Alignment with Reference Answers}

Semantic Alignment requires the comparison of model-generated responses to reference answers in both the lexical and semantic levels. Automated measures are used to evaluate superficial overlap and deeper meaning correspondence. These metrics are ROUGE -1, ROUGE-2, ROUGE-L, Jaccard Similarity and Cosine Similarity.  These measures capture complementary aspects of model behavior and allow separation of token-level correspondence from semantic proximity.\\
There is a significant difference between models based on the metric used. For instance, Gemini scores highest on lexical overlap measures with average ROUGE-1, ROUGE-2, and ROUGE-L scores of approximately 0.56 0.22 0.16 and the largest Jaccard Similarity 0.21. These findings demonstrate a high level of superficial similarity to reference response, as they suggest that Gemini is capable of reproducing phrasing and token structure. Nevertheless, lexical similarity does not necessarily mean closer semantic correspondence.\\
The closest similarity is the semantic similarity, which is evaluated by the cosine similarity and has the highest value of Claude with an average of approximately 0.68 and a maximum of 0.80. Cosine similarity scores of GPT and Perplexity are more similar implying that the two models reflect the underlying meaning of the reference responses even in cases where the overlap between lexicons is reduced. In comparison with lexical metrics, which are more diverse (e.g., ROUGE -1 scores have a range between 0.16 and 0.56), the cosine similarity scores fall in a smaller range (around 0.63 to 0.71), meaning that the semantic interpretation of the models is more consistent.\\

\begin{figure}[H]
  \centering
  \includegraphics[width=0.7\textwidth]{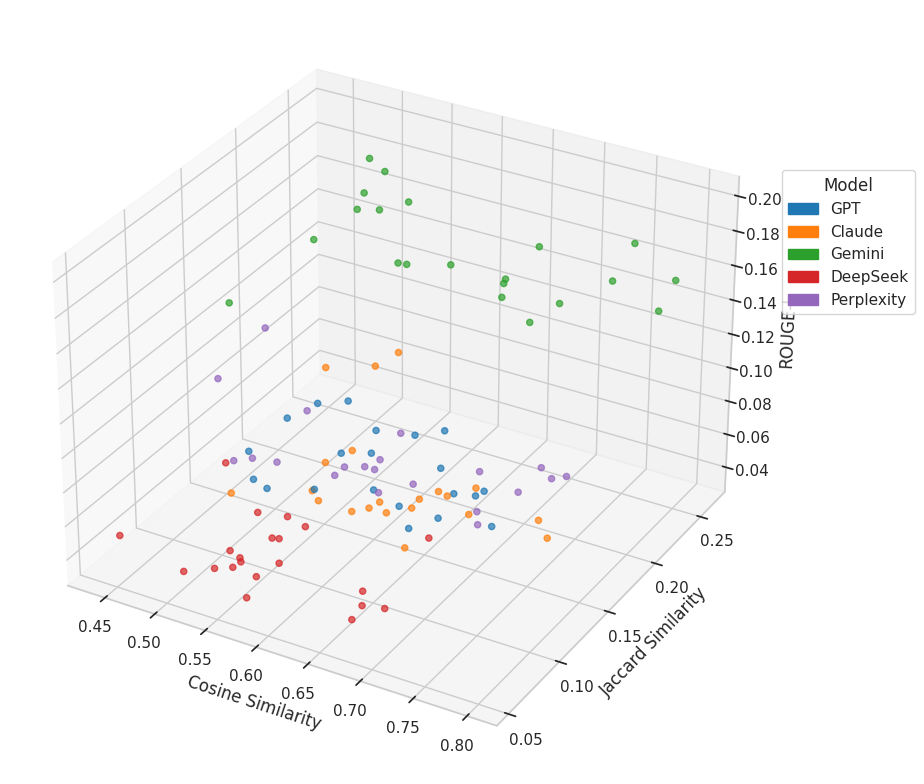}
  \caption{Joint distribution of cosine similarity, Jaccard similarity, and ROUGE-L across models.}
  \label{fig:3D}
\end{figure}

These patterns are illustrated in  Figure~\ref{fig:3D} as the visualization of lexical overlap and semantic alignment of models is jointly used. The figure shows that models that are similar in lexicon do not necessarily fall at the same point in semantic proximity, which supports the importance of making the distinction between surface-level fidelity and conceptual compatibility. On the whole, the findings have shown that both lexical fidelity and semantic alignment are different aspects of model behavior. In the context of financial text analysis, lexical overlap can be exaggerated as the measure of the interpretative quality in the situation when the understanding of semantics and the consistency of meaning and context are the main aims.

\subsection{Consistency, Stability, and Cross-Model Agreement}

This task assesses the consistency of the behavior of the models with respect to various prompts, companies, and time. It is more concerned with the correspondence between models and the stability of the correspondence of models with time, as opposed to the accuracy of individual outputs. With behavioral diagnostics, we consider whether the behavior of the models in the interpretation of the same financial data is similar and whether the interpretation remains similar as the disclosure content varies.

Pairwise analysis of semantic similarity demonstrates that GPT and Claude are in agreement through most company datasets with a cosine similarity of over 0.80 in most instances (e.g. 0.84 in Apple and 0.87 in Microsoft). This shows that both models have very close semantic representations in the interpretation of financial disclosures. Model pairs to which Gemini or DeepSeek are added, by comparison, are less predictive, with cosine similarities falling to approximately 0.69 -0.72 in certain cases - particularly with companies and years which disclose more or whose disclosures are more changing. These reduced scores demonstrate higher variability in the encoding and the priorities of those models to financial information.

Temporal analysis also helps determine that the cross-model agreement is conditional upon the alteration of disclosure content. Microsoft reports, e.g., demonstrate no significant change over time, and the mean cosine similarities are 0.85 with a poor variance. However, other firms (such as Amazon) are more volatile: mean similarities are close to 0.71 and standard deviations are as high as approximately 0.078 in more recent filings. This greater variance indicates that the less developed or less straightforward disclosures elicit model variance.

\begin{figure}[H]
  \centering
  \includegraphics[width=0.7\textwidth]{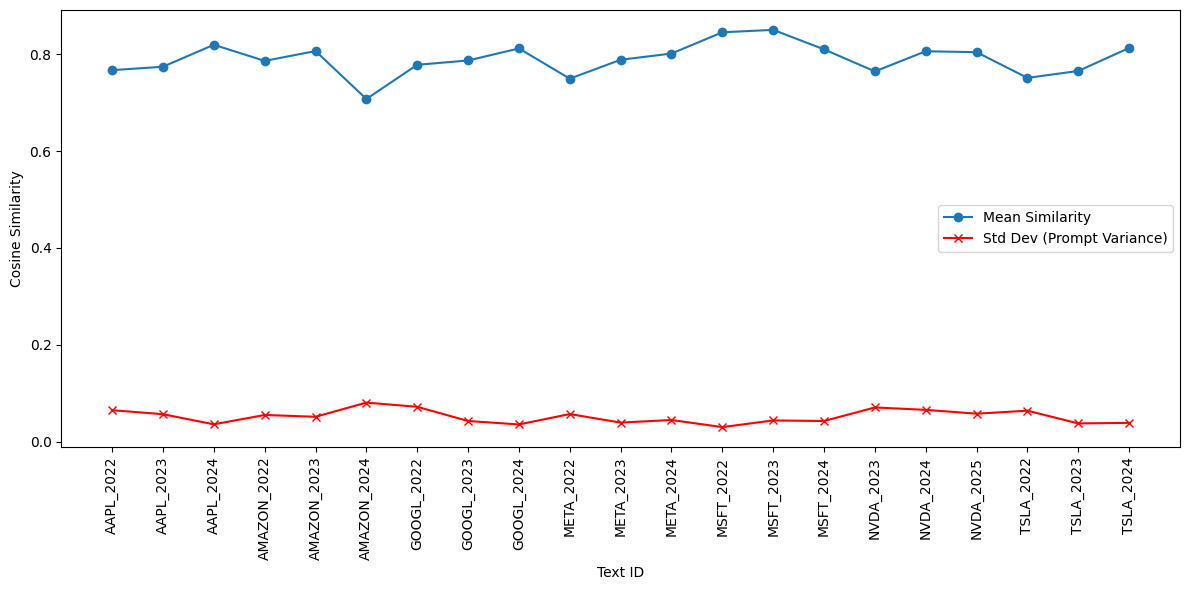}
  \caption{Mean and variance of cosine similarity across firms and years.}
  \label{fig:cosine_variance}
\end{figure}

These time and cross-model trends appear in Figure~\ref{fig:cosine_variance}. The number gives the average and standard deviation of the similarity of the cosine between the firms over years. A high average agreement and low variance means there is increased interpretive stability. A lower mean similarity and a larger variance, in contrast, are sensitive to disclosure content changes. In general, the results note that consistency and stability are dissimilar dimensions of model behavior. Lexical overlap and semantic similarity cannot in themselves allow them to be inferred. Those models, which exhibit more prompts and time agreement, can have more reliability in terms of their application in the applications which need a consistent interpretation. Conversely, when they are more varied, they may be less applicable to repeated or longitudinal financial analysis.

Combined, the findings of the three assessment activities indicate that the behavior of LLMs in the context of processing financial text is predetermined by specific and even contradictory factors. In determining the disclosure in 10-K reports, qualitative tests bring out the balance of relevance, completeness, conciseness and factual accuracy models. Computerized metrics show that in surface lexical overlap and semantic alignment on deeper levels complementary, though not equivalent, features of model output are measured. The behavioral diagnostics depict that model consistency and stability over time differ significantly in meaningful ways among different firms and reporting periods.

In tasks, models with better semantic convergence and reduced response dispersion are more likely to give more consistent readings of financial disclosures, particularly where repeated or longitudinal studies are to be undertaken. On the contrary, high lexical similarity in itself does not also show a stronger interpretive alignment, which highlights the shortcomings of surface-level measures of financial language understanding. The variations in firms and years indicate that model behavior is sensitive to the timely formulation as well as the modification in the content of disclosure and hence emphasizes the considerations of constant review due to the changes in the language of financial reporting.

In general, all these results indicate that no particular evaluation dimension can be used to describe the model appropriateness in financial text analysis. Qualitative judgment, lexical and semantic alignment measures and behavioral consistency are measures that capture different aspects of model behavior. The interrelation between these views through a multidimensional assessment framework can offer a more informative foundation into how the interpretation of financial disclosure by LLMs is as well as in what contexts their application can be best utilized in practical financial environments.

\section{Limitations}

There are a number of methodological limitations to be observed in the interpretation of the results. The disclosures made by the firms we used are unusually detailed and of high quality and are not necessarily reflective of the variability observed by smaller issuers, by non-technology industries, or by other regulatory reports, such as MD\&A or 10-Q reports. The small sample size restricts the statistical applicability of the results, making the study an exploratory study of model behavior in controlled disclosure conditions instead of an indicator of a population point.

Second, the human assessment was done by five trained annotators who had engineering and NLP background, and not domain-specialized financial analysts. Although the annotators were calibrated using a standardized rubric and advised to pay attention to factuality and relevancy in relation to the given text, their judgments might be different as compared to people who are professional in using financial disclosures. The measures of inter-rater reliability reduce, although fail to remove, the potential source of systematic bias due to annotator background.

Third, the experimental design has set in advance one prompting scheme, deterministic decoding parameters when available, and a uniform truncation scheme to confer the smallest model context window. Such decisions enhance reproducibility at the expense of reducing the space of evaluation. Other prompting methods, sampling schemes or document-segmentation schemes might produce varying comparative behaviours of models.

Lastly, the research takes a one time performance of models. Proprietary LLMs are developed quickly, and parameters change frequently, as does intervention, and interface. As a result, the reported results can be interpreted as time-constrained approximations of model behaviour as opposed to performance guarantees.

Taken together, the restrictions indicate future research opportunities such as using larger and more diverse regulatory corpora and domain-specialist assessment, larger prompt families, and tracking model version drift over time in financial-text contexts.

\section{Conclusion}\label{sec2}

LLMs are becoming an inseparable part of the current natural language processing systems, particularly in areas where it is important to comprehend and synthesize intricate texts. In finance, these models have become common to complete tasks like document summarization, narrative analysis and decision support. The applications provoke some serious doubts regarding reliability, transparency, and consistency of behavior. Despite the high capabilities of LLMs, they are hard to interpret, and their results might be highly divergent with prompts or contexts and model architectures. This may have actual implications in high stakes environments because of such differences in high-stakes settings, explained by \cite{mohsin_explainable_2025}.

The present study was a controlled and multi-faceted assessment of five transformer-based LLMs on Business section excerpts of 10-K filings. Human assessment of model behavior was done by evaluation based on human evaluation, automated similarity measures and behavioral diagnostics evaluating not just surface-level performance but also consistency and similarity of response across prompts. Through these supplementary lenses, the analysis provides a systematized perception of the behaviour of the various models in the event of the standardized financial disclosure of inputs, as opposed to a univariate evaluation of the aggregate performance.

In many of the qualitative criteria in the evaluated dimensions, GPT scored higher in the human evaluation protocol on their average scores. The other models demonstrated their strengths in certain areas, e.g. lexical overlap or conciseness in the responses. Nevertheless, inter-rater reliability was low in all the qualitative dimensions, which means that experts are quite different in their evaluations. There is no one model that will always prevail within the standards of evaluation. The above results imply that the observed apparent performance disparities must be seen as relative tendencies in the conditions of the test, rather than absolute examples of the superiority or the general validity.

In a wider context, the findings indicate that unrefined performance indicators cannot be used to determine the appropriateness of LLMs to be applied in money-related contexts. When these systems aid analysis or decision-making, behavioral stability, response to timely changes, as well as the interpretability of a model output, are also essential. The perceived inconsistency among models and the methods of evaluations is a reminder that a careful deployment is necessary and that the frameworks of evaluations should explicitly take into consideration the uncertainty and disagreement in the way humans assess.

Subsequent studies would be improved by using more extensive and more varied financial corpora, using evaluators with domain-specific expertise, and examining other prompting and decoding methods. Further development of causal interpretability, model alignment, and explanation control will also play a significant role in converting the capabilities of LLM to reliable financial uses. This paper will add to a growing literature on this topic by focusing on comparative behavior, reliability, and transparency instead of absolute rankings to make responsible use and evaluation of LLMs in finance.

\end{document}